# Dynamic Cat Swarm Optimization Algorithm for Backboard Wiring Problem

**Aram Ahmed [1,2], Tarik Rashid [3] and Soran Saeed [1]**

[1] Department of Information Technology, Technical College of Informatics, Sulaimani Polytechnic University, Sulaimanyah, Iraq.
[2] International academic office, Kurdistan Institution for Strategic Studies and Scientific Research, Sulaimanyah, Iraq.
[3] Department of Computer Science and Engineering, School of Science and Engineering, University of Kurdistan Hewler, Erbil, Iraq.
\* Correspondence: aramahmed@kissr.edu.krd;

## Abstract

This paper presents a powerful swarm intelligence meta-heuristic optimization algorithm called Dynamic Cat Swarm Optimization. The formulation is through modifying the existing Cat Swarm Optimization. The original Cat Swarm Optimization suffers from the shortcoming of "premature convergence", which is the possibility of entrapment in local optima which usually happens due to the off-balance between exploration and exploitation phases. Therefore, the proposed algorithm suggests a new method to provide a proper balance between these phases by modifying the selection scheme and the seeking mode of the algorithm. To evaluate the performance of the proposed algorithm, 23 classical test functions, 10 modern test functions (CEC 2019) and a real world scenario are used. In addition, the Dimension-wise diversity metric is used to measure the percentage of the exploration and exploitation phases. The optimization results show the effectiveness of the proposed algorithm, which ranks first compared to several well-known algorithms available in the literature. Furthermore, statistical methods and graphs are also used to further confirm the outperformance of the algorithm. Finally, the conclusion as well as future directions to further improve the algorithm are discussed.

## 1. Introduction

The aim of optimization is to identify the best solution for a specific problem among many alternative solutions. The objective can be minimization, such as minimizing cost or time, or it can be maximization, such as maximizing profit or production. There are two main families of optimization algorithms, namely exact and approximate algorithms [1]. In the first one, some specific rules are used in which the same inputs will give the same outputs all the time. However, in the second type, because it always has some randomness and depends on some probabilistic rules, the same inputs would not necessarily give the same outputs [2]. Metaheuristic algorithms are of approximate types. They are referred as a high-level scheme that directs and changes other heuristics in order to generate solutions further than those solutions which are typically achieved in the pursuit of local optima [3]. It is proven that when the number of possible solutions is large enough that makes it infeasible for the exact algorithms to be used, metaheuristic algorithms come to play their important role by providing fairly decent solutions during an acceptable time[4].

Swarm intelligence algorithms are of metaheuristic types, in which multiple agents move in the search space. These agents are decentralized and self-organized, which somehow communicate and work together to find the optimal solution [4]. In general, these agents perform the search in two phases, namely exploration and exploitation. Exploration means jumping out to search for new regions on a global scale, while exploitation means focusing on those regions that have already been searched to find better solutions. Over-exploration of an algorithm will lead to diversification of agents and it is unlikely to reach the near-optimum solutions. On the other hand, over-exploitation of an algorithm will increase the chance of falling into local optima and not being able to find the global optimum. Therefore, having a proper balance between these two phases is very critical in any metaheuristic algorithm [5].

Cat Swarm Optimization (CSO) is a metaheuristic algorithm, which is inspired by the resting and hunting behavior of cats. The algorithm has the limitation of being trapped into local optima. Therefore, many modified versions have been developed to overcome this issue. For example, Pappula et al. used a normal mutation technique [6]; Nie et al. adopted quantum mechanics and tent map techniques [7], Orouskhani et al. used an inertia value to the velocity equation [8], etc. A relevant detailed survey can be found in [9]. This paper proposes a Dynamic Cat Swarm Optimization (DCSO) algorithm, in which the main contribution is to provide a proper balance between the global and local searches and hence avoid the local optima trap. In order to achieve this, the selection scheme and the seeking mode of the algorithm are modified. Regarding the selection scheme, MR parameter was replaced with two new adaptive parameters and the selection process was changed to be fitness dependent instead of random selection. As for the seeking mode modification, greedy method was used instead of roulette wheel method and SRD parameter was removed as well. This idea was first introduced in Simulated Annealing algorithm by Kirkpatrick et al. [10]. Therefore, DCSO algorithm exploits the idea to propose a novel adaptive method and improve the original CSO algorithm.

The rest of the paper is organized as follows: Section 2 presents the motivation and novelty of this study. Section 3 describes the Backboard Wiring Problem. Section 4 discusses the methodology used to formulate the original CSO algorithm as well as the proposed DCSO algorithm. Section 5 presents the results and discussions, where the proposed algorithm is evaluated against Chimp Optimization Algorithm (ChOA) [11], the original CSO algorithm (CSO) [12] and Differential Evolution (DE) [13]. Finally, section 6 concludes this study and provides future directions.

## 2. Motivation and Novelty

The tracing and seeking modes of CSO algorithm are playing the role of exploitation and exploration phases of the algorithm respectively. These two modes are controlled by a parameter called Mixture Ratio (MR). This parameter has two weaknesses that critically affect the performance of the algorithm, with the following details:

1. It is a static parameter and its value is unchangeable throughout the iterations. This prevents the algorithm from assigning the right number of cats to the exploration or exploitation phases at different stages of optimization process.
2. The cats assigned to each mode, are selected randomly. This also causes confusion to the agents and misleads them to defective areas. Suppose there is a cat in a promising area, this cat should be assigned to the seeking mode for the next iterations. However, CSO algorithm might randomly assign the cat to the tracing mode, which takes the cat away from the promising area and transfers it towards a local optimum.

DCSO algorithm proposes the following modifications to tackle these issues and balance between exploration and exploitation phases:

1. The MR parameter is replaced with TCN and SCN parameters, which are adaptive and their values change according to the number of iterations. These two parameters form a dynamic selection scheme, where at the beginning of iterations two cats are assigned to the tracing mode and the remaining cats are assigned to the seeking mode. Then progressively and according to the number of iterations, the number of tracing cats increase and the number of seeking cats decrease i.e., TCN and SCN parameters are inversely proportional. Thus, at the last iteration all cats are in the tracing mode, and are moving towards the global optimum. This means that at the early iterations, the algorithm performs global search to find the promising areas. Then, the algorithm makes a smooth and gradual transition toward local search.

2. The proposed algorithm, in each iteration, sorts the cats according to their fitness costs. This sorting is used to distinguish between the cats with higher fitness cost and those with lower fitness costs. Therefore, cats with lower fitness costs are always assigned into the seeking mode to further explore the promising area, and not to leave it prematurely. Meanwhile, cats with higher fitness costs are assigned to tracing mode because it is preferable for these cats to leave their defective areas and move towards the global best.

3. The proposed algorithm also has two modifications in the seeking mode. First, greedy method is used instead of roulette wheel method. Second, SRD parameter is removed and only a random ratio is used to mutate the positions. As the name implies, greedy method always selects the best candidates as its next position. On the other hand, the roulette wheel method gives the chance to the other solutions to be selected as well. Therefore, the greedy method is in favor of seeking phase as it allows the agents to move to better areas only.

## 3. DCSO algorithm for Quadratic Assignment Problem: Backboard Wiring Problem

This section is dedicated to discuss the concept and mathematical details of the backboard wiring problem. Moreover, it explains how the DCSO algorithm is applied on the problem.

### 3.1 Problem description

Backboard Wiring Problem is an optimization problem, which considers the placement of electronic elements on a computer backboard so as to minimize the total length of wires required to connect them. Solving such problem plays an important role in enhancing the speed and efficiency of the system. In the example of Steinberg 34 components with a total of 2625 interconnections are involved to be located on 36 positions in a backboard. Two dummy components are also added to equalize the number of components and locations [14]. Figure 1 is the geometry of the backboard wiring problem where a possible permutation for a dataset instance is presented [15].

| P1 | P2 | P3 | P4 | P5 | P6 | P7 | P8 | P9 |
|----|----|----|----|----|----|----|----|----|
| 35 | 31 | 30 | 29 | 28 | 1  | 15 | 9  | 16 |
| P10| P11| P12| P13| P14| P15| P16| P17| P18|
| 33 | 34 | 32 | 19 | 20 | 7  | 10 | 18 | 17 |
| P19| P20| P21| P22| P23| P24| P25| P26| P27|
| 26 | 25 | 23 | 14 | 12 | 13 | 4  | 8  | 2  |
| P28| P29| P30| P31| P32| P33| P34| P35| P36|
| 24 | 22 | 21 | 27 | 11 | 6  | 5  | 3  | 36 |

**Figure 1.** A solution for backboard wiring problem (dataset: ste36b) yielding the cost value of 15852

### 3.2 Problem formulation

Mathematically, this problem can be formulated as the Quadratic Assignment Problem (QAP). Practically, QAP is considered as one of the most complicated combinatorial optimization problems. It was first introduced by Koopmans and Beckmann and it has many other real-world applications, such as hospital layout facility, airport gate assignment, dartboard design, etc. [16]. Given n elements, n locations, and two n x n matrices, the flow matrix $A = (a_{ik})$, and the distance matrix $D = (d_{jl})$. Here, $a_{ik}$ denotes the number of wires, which connect elements $i$ and $k$, and $d_{jl}$ denotes the distance between locations $j$ and $l$ on the backboard. Therefore, the formulation of Steinberg Wiring Problem (SWP) can be written as:

$$\min \sum_{i=1}^{n} \sum_{j=1}^{n} \sum_{k=1}^{n} \sum_{l=1}^{n} a_{ik}\, d_{jl}\, x_{ij}\, x_{kl}$$

Subject to

$$\sum_{j=1}^{n} x_{ij} = 1,$$

$$\sum_{i=1}^{n} x_{ij} = 1,$$

(1)

Where

$$x_{ij} = \begin{cases} 1, & \text{if facility } i \text{ is assigned to location } j \\ 0, & \text{otherwise} \end{cases}$$

### 3.3 Problem constraints

As stated in the previous subsection, $x_{ij}$ matrix would be 1 if element $i$ is assigned to location $j$ on the backboard and it will be 0 otherwise. The constraints guarantee that each element $i$ is allocated to exactly one location $j$ and each location $j$ contains one element $i$ merely.

### 3.4 Problem objectives

The problem involves assigning $n$ elements on $n$ locations on a backboard while minimizing the assignment cost value. This assignment is denoted by a permutation, which specifies the location of each element [17]. Therefore, the below steps are followed in order to apply the DCSO algorithm on the problem:

1. The algorithm generates a population of cats (solutions) in each iteration.
2. The position values of each cat are sorted in ascending/descending order.
3. The indices of these sorted values are obtained in order to create a permutation.
4. The achieved permutation will be used in equation (1) to calculate the cost value.
5. After calculating the cost values for all the permutations, the one with the least cost value is selected as the best permutation.

For example, let's assume that lower and upper bounds are equal to 0 and 1 respectively. Let's also assume that the position of a four-dimensional Agent is like: (0.12, 0.74, 0.01, and 0.46). Therefore, by taking the indices of their sorted values, a permutation like (2, 4, 1, and 3) is obtained. The sets of permutations of all agents are then used in equation (1) to calculate their fitness values. Next, the best permutation is chosen which has the least cost value.

## 4. Cat Swarm Optimization

This section consists of two main parts, where both of the original Cat Swarm Optimization (CSO) Algorithm and the proposed Dynamic Cat Swarm Optimization (DCSO) Algorithm are presented respectively.

### 4.1 The Original Cat Swarm Optimization

Cat Swarm Optimization (CSO) was originally developed by Chu and Tasi in 2006 [18]. It is inspired by two main behaviors of cats, which are resting and hunting. Accordingly, the algorithm consists of two modes, namely seeking and tracing modes. Each cat denotes a solution set, which has its own position, a cost value, and a flag. The position is comprised of M dimensions, where each dimension has its own velocity. The cost value reveals how well a solution set (cat) is. Finally, the flag is to specify whether the cat is in seeking mode or tracing mode. In each iteration, the best cat is identified, which represents the best solution found so far.

*4.1.1 General Steps for CSO Algorithm*

The algorithm takes the following steps while looking for the optimal solution:

1. Identify the upper and lower bounds for the solution sets.
2. Randomly create N cats and scatter them into the M dimensional space. Each cat should have a random velocity value within the predefined velocity limits.
3. According to MR value, the cats are randomly assigned to seeking mode or tracing mode. MR parameter is a mixture ratio, which is selected in the range of [0, 1]; For instance, if N was equal to 10 and MR was equal to 0.2, then 2 cats will be randomly selected to go through the tracing mode and the remaining 8 cats will go through the seeking mode.
4. Evaluate the cost value for all cats and then the best cat of the current generation will be identified.
5. Move the cats according to their flags. If a cat was assigned to the seeking mode, apply the cat to the seeking mode process, if not apply it to the tracing mode process.
6. For the next iteration, re-pick the cats according to MR and set them to go through either seeking or tracing modes.
7. Check the termination condition, if satisfied; terminate the program; otherwise, repeat Step 3 to Step 6.

*4.1.2 Seeking mode*

This mode mimics the resting behavior of cats and has four vital parameters, which are seeking memory pool (SMP), seeking range of the selected dimension (SRD), counts of dimension to change (CDC) and self position consideration (SPC).

SMP identifies the number of copies of a cat (candidate positions) to be produced in the seeking mode. For instance, if SMP was set to 5, then for each cat in the seeking mode, 5 random copies will be generated in which the cat chooses one of them as its next position. The way these random copies are generated depends on CDC and SPC. CDC parameter is in the range of [0, 1] and specifies how many dimensions to be modified. For instance, if CDC was set to 0.8 and the number of dimensions in the search space was 10, then for each cat 8 dimensions will be modified and the rest will stay constant. SRD is a mutative ratio for those dimensions, which are selected by CDC. Lastly, SPC is a Boolean valued parameter, which specifies whether the current position of the cat will be selected as one of the copies of SMP or not. For instance if SPC was set to true and SMP was set to 5, then only 4 random copies will be generated and the current position will be the 5th candidate position. The steps of this mode are as follows:

1. Make j copies of the current position of a cat, where j = SMP. if SPC is true, then j=(SMP-1) and keep the current position as one of the candidates.
2. For each copy, according to CDC choose some random dimensions to be changed. Then, randomly add or subtract SRD values from the existing positions as shown in equation (2):

$$X_{j,d} = (1 \pm \text{rand} * \text{SRD}) * X_{j,d} \quad (2)$$

Where $X_{j,d}$ is the cat's position; $j$ and $d$ represent the number of copies and dimensions for the cat respectively;

3. Calculate the cost value for the candidate positions.
4. Using the roulettewheel method, calculate the selecting probability of each candidate point according to equation (3). Therefore, Candidate points with better fitness costs have more chances to be selected. However, if all fitness costs were the same, then the selecting probability of each candidate point would be 1.

$$Pi = \frac{|FS_i - FS_b|}{FS_{max} - FS_{min}}, \quad \text{where } 0 < i < j \quad (3)$$

If the objective is minimization then FSb = FSmax, otherwise FSb = FSmin.

*4.1.3 Tracing mode*

This mode mimics the hunting behavior of cats and the steps are as follows:

1. Update the velocities for all dimensions according to equation (4):

$$V_{k,d} = V_{k,d} + c_1 * \text{rand} * (X_{best,d} - X_{k,d}) \quad (4)$$

Where $V_{k,d}$ is the velocity for the $k^{th}$ cat in the $d^{th}$ dimension; $X_{best,d}$ is the position of the cat with the best fitness cost; $X_{k,d}$ is the current position of the $k^{th}$ cat in the $d^{th}$ dimension. $c_1$ is a constant and $rand$ is a single uniformly distributed random number in the range of [0, 1].

2. Set the new velocity value to the limits if it out-ranged the bounds of velocity.
3. Update the position of $k^{th}$ cat according to equation (5):

$$X_{k,d} = X_{k,d} + V_{k,d} \quad (5)$$

Where $X_{k,d}$ is the position of $k^{th}$ cat in the $d^{th}$ dimension.

## 4.2 The proposed Dynamic Cat Swarm Optimization

The MR parameter in the original CSO specifies what percentage of the population to go through the seeking or tracing modes. It is a static parameter, which has a fixed value throughout the entire iterations of the algorithm. DCSO reformulates this selection scheme by equations (6) and (7):

1. Tracing Cat Number (TCN): specify the number of cats, in $i^{th}$ iteration, that are entering the tracing mode and it is defined according to equation (6):

$$\text{TCN} = \left\lfloor \frac{i * N}{\text{MaxIter}} \right\rfloor \begin{cases} \text{TCN} = 2 & \text{if TCN} \leq 2 \\ \text{TCN} = \text{TCN} & \text{if TCN} > 2 \end{cases} \quad (6)$$

Where $i$ is the current number of iteration; $MaxIter$ is maximum iteration; $N$ is the population size.

2. Seeking Cat Number (SCN): specify the number of cats, in $i^{th}$ iteration, that are entering the Seeking modes and it is defined according to equation (7):

$$\text{SCN} = N - \text{TCN} \quad (7)$$

$TCN$ and $SCN$ are inversely proportional. Together, they form a selection scheme, in which at the beginning of iterations, only two cats are entering the tracing mode and the rest of them are entering the seeking mode. Then, gradually and according to the number of iterations, cats are removed from the seeking mode and sent to the tracing mode. Hence, by the end of iterations, all cats are in the tracing mode to move towards the best solutions found so far. Accordingly, Figure 2 shows how the agents are chasing the global best and how the number of seeking cats (squares) decreases in an inverse proportional manner to the number of tracing cats (triangles) throughout iterations.

*4.2.1 General Steps for DCSO Algorithm*

The algorithm takes the following steps while looking for optimal solutions:

1. Initialize the cat population.
2. Calculate the fitness function for all cats.
3. According to their fitness costs, sort the cats from best to worst and identify the best solution of the current generation.
4. Calculate TCN and SCN (equations 6 and 7).
5. Select as many cats as SCN value from the best cats of the population and Send them to the seeking mode.
6. Send the remaining cats into the tracing mode.
7. Combine and mix all tracing and seeking cats together.
8. If any cat dimension is out of the boundary, it is equal to the limits, as in equation (8):

$$X_{k,d} = \begin{cases} L_d & \text{if } X_{k,d} < L_d \\ U_d & \text{if } X_{k,d} > U_d \end{cases} \quad (8)$$

Where $X_{k,d}$ is the position of $k^{th}$ cat in the $d^{th}$ dimension; $L_d$ and $U_d$ are lower and upper bounds for $d^{th}$ dimension of the search space.

9. Calculate the fitness cost for all cats and update the global best.
10. Check if the termination condition is satisfied, then terminate the program. Otherwise, repeat Step 3 to Step 9.

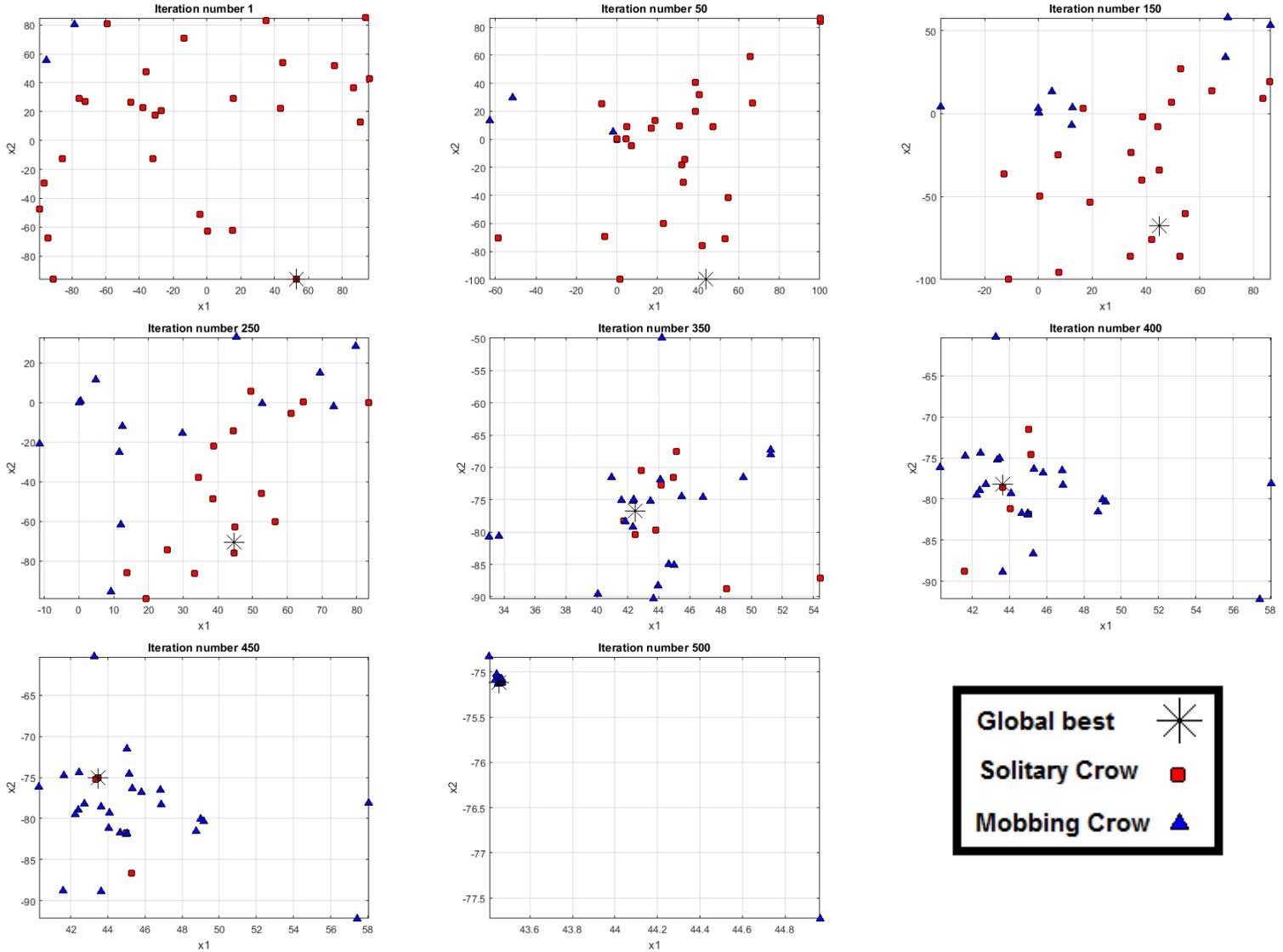

**Figure 2.** Number of seeking and tracing cats in different iterations. As number of iterations increases, number of tracing cats (exploitation phase) increases and number of seeking cats (exploration phase) decreases.

*4.2.2 Tracing mode*

This mode is similar to the original CSO algorithm. However, in DCSO algorithm inertia weight, which is a linear time varying parameter, is added to the velocity equation. This parameter is used repeatedly in many metaheuristic algorithms to linearly decrease the velocity value and hence balance between the exploration and exploitation phases [8]. So, the algorithm takes the following steps in this mode:

1. The velocities for each dimension are updated according to equation (9):

$$V_{k,d} = w * V_{k,d} + c_1 * \text{rand} * (X_{best,d} - X_{k,d}) \qquad (9)$$

2. The positions of cats are updated according to equation (10):

$$X_{k,d} = X_{k,d} + V_{k,d} \qquad (10)$$

*4.2.3 Seeking mode*

In this mode, DCSO algorithm has two main modifications: firstly, greedy method is used instead of the roulettwheel method. Thus, instead of selecting the candidate positions based on probability, best candidate position is always selected. Secondly, the SRD parameter is removed and only a random value in the range of (0, 1) is used. So, the steps of this mode are as follows:

1. For each cat that comes into this mode, make j copies of its current position, where j = SMP.
2. For each copy, in a random permutation manner, select as many dimensions as CDC value.
3. Calculate the copies according to equation (11):

$$X_{j,d} = (1 \pm \text{rand}) * X_{j,d} \qquad (11)$$

4. Fitness cost for all copies are calculated and using the greedy method the best copy is selected to be the next position for the cat.

Figure 3 is for illustrative purposes; it shows how a cat moves in seeking or tracing modes. Additionally, Figure 4 is the flowchart of DCSO algorithm, which presents the sequence steps of the algorithm.

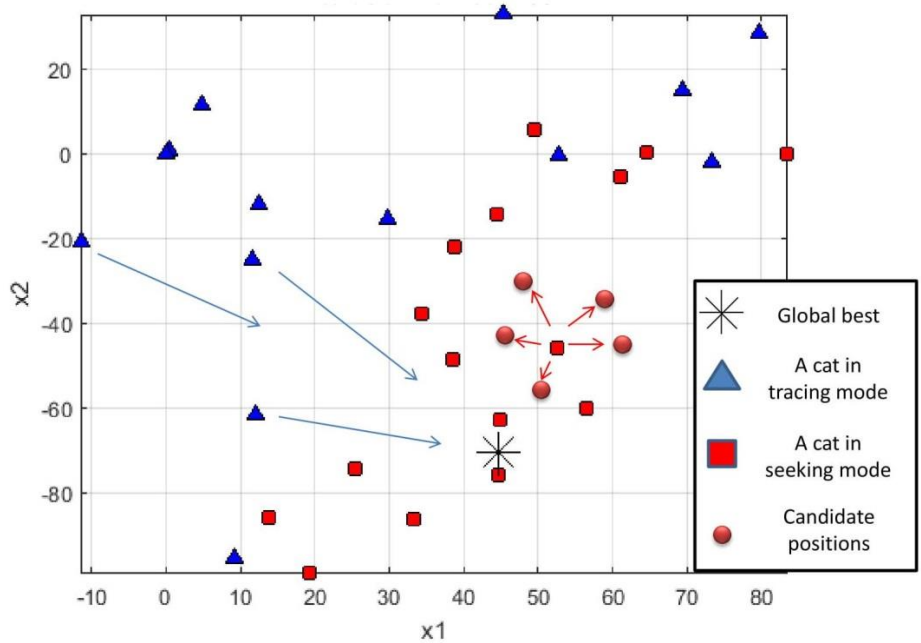

**Figure 3.** The movements of cats in DCSO algorithm: cats in tracing mode move towards the global best, which represents the exploitation phase. Cats in seeking mode move towards a candidate position, which represents the exploration phase.

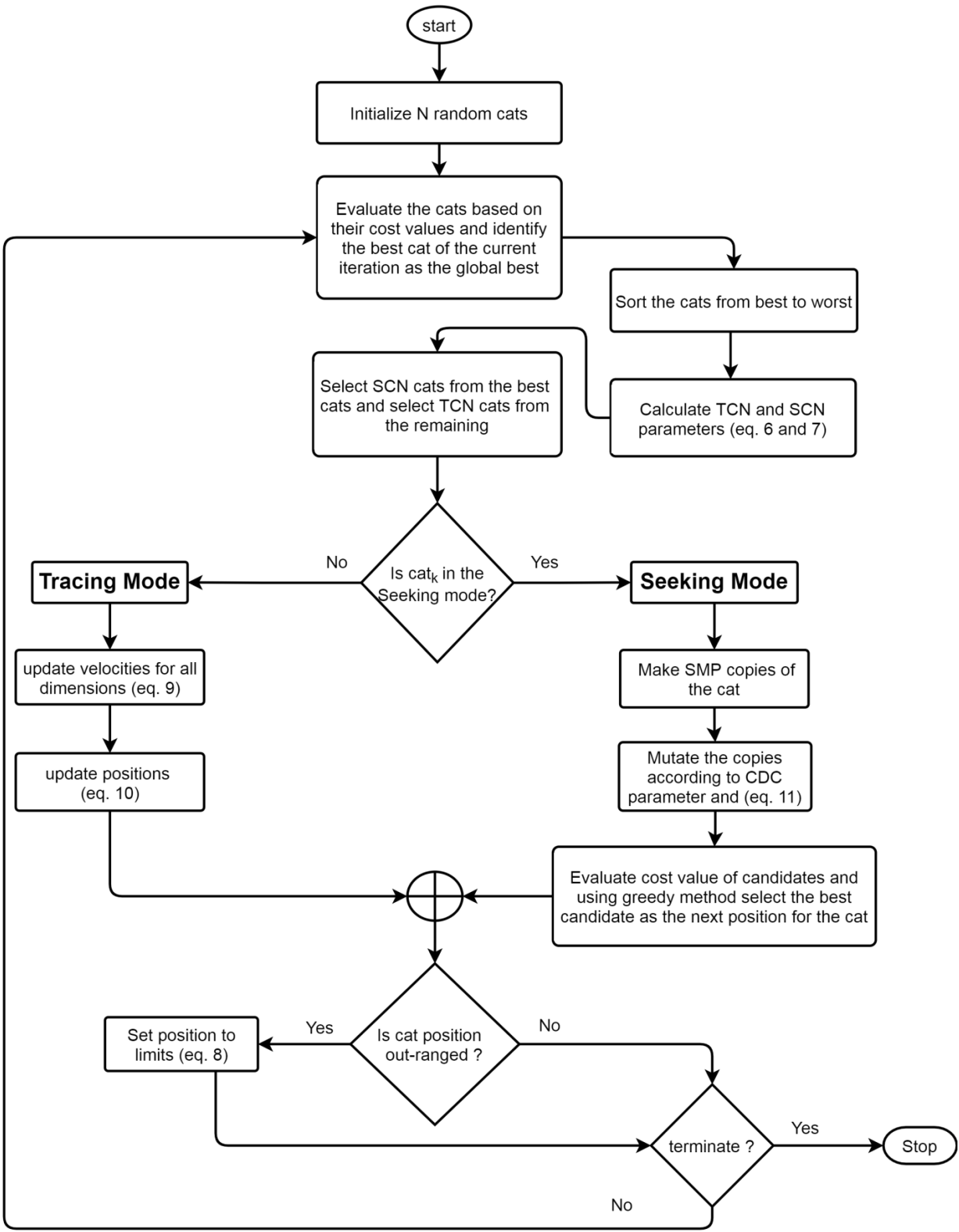

**Figure 4.** The flowchart of DCSO algorithm

## 5. Results and discussions

In this section, the performance of the proposed algorithm is evaluated. Therefore, the algorithm was benchmarked on thirty-three test functions and a real-world scenario. Besides, the time and space complexity of the algorithm as well as the dimension-wise diversity measurement are presented. Figure (5) presents the general performance evaluation framework for the algorithm.

The test functions are classified into two groups, which are classical and modern test functions.

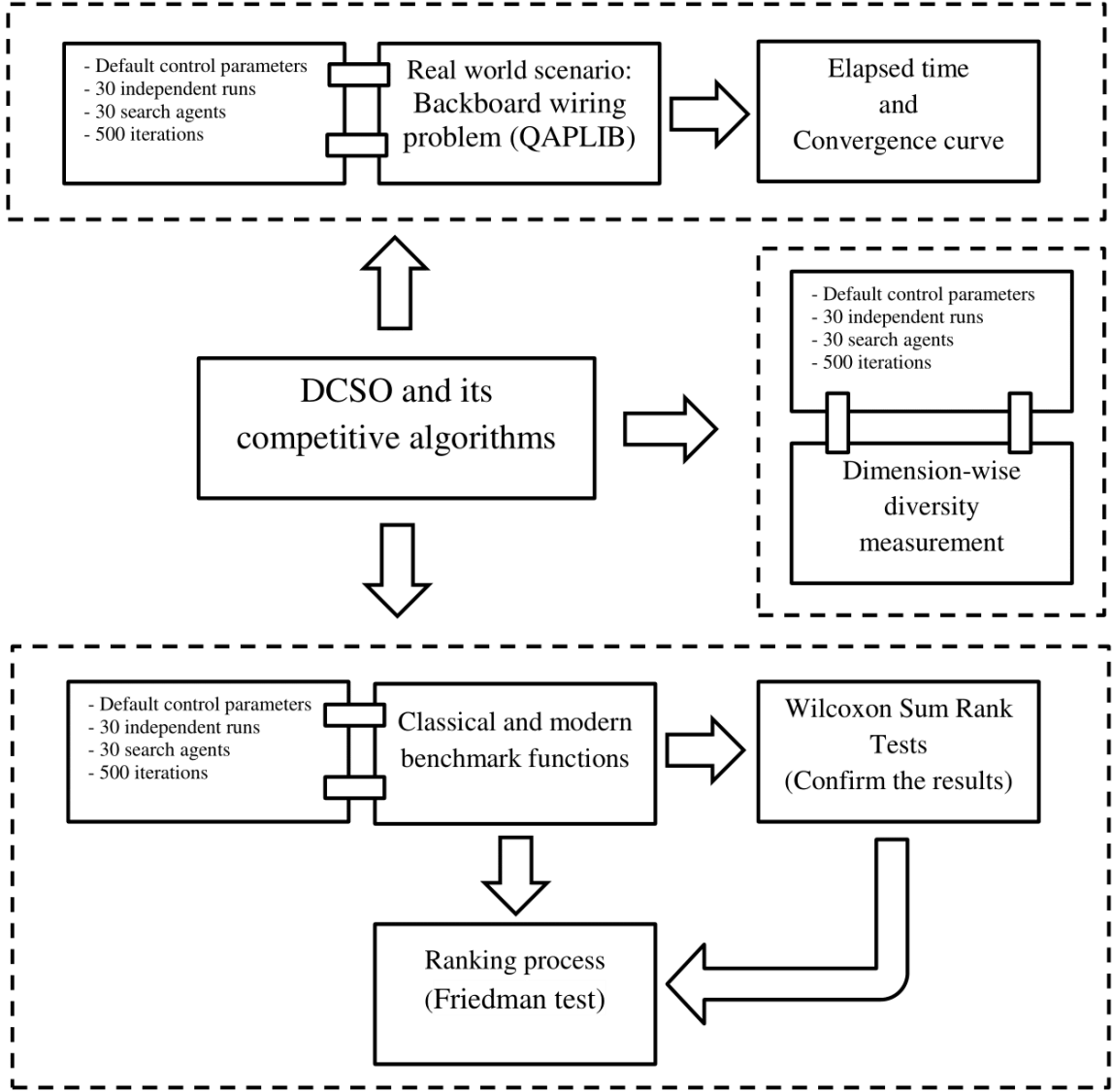

**Figure 5.** The general performance evaluation framework for the proposed algorithm

### 5.1 Classical Test Functions

This group contains the unimodal and multimodal test functions, see Table 6 in the appendix for the detailed description of these functions. F1 to F7 are of the unimodal type and they are used to test the exploitation ability of an algorithm since they have one single optimum. However, F8 to F23 are Multimodal test functions, which have multiple local optima and one of them is the global optimum. Therefore, these types of functions are used to test the exploration ability of an algorithm.

### 5.2 Modern Test Functions (CEC 2019)

These test functions are also known as composed test functions. They are shifted, rotated, hybridized or expanded versions of other test functions. They are considered to be the most challenging benchmark functions and they are used to test how well an algorithm can balance between the exploration and exploitation phases. For this work, CEC 2019 special session, also known as "The 100-Digit Challenge", is used as composed functions [19], (see Table 7 in the appendix).

The comparison results of DCSO algorithm for unimodal, multimodal and CEC 2019 benchmark functions are presented in Tables 1 and 2 in the form of mean and standard deviations. For each benchmark function, the algorithm is executed for 30 independent runs. For each run, the number of search agents and iterations is equal to 30 and 500 respectively. Furthermore, the results are compared with three well-known algorithms, namely Chimp Optimization Algorithm (ChOA) [11], Cat Swarm Optimization algorithm (CSO) [12] and Differential Evolution (DE) [13]. Parameter settings for these algorithms are presented in Table 8 in the appendix. Furthermore, by looking at Tables 1 and 2, it is clear that DCSO algorithm yields competitive results.

**Table 1.** Comparison results between DCSO and the Selected Algorithms for Classical Test Functions in terms of Average and Standard Deviation

| F | DCSO | | ChOA | | CSO | | DE | |
|---|---|---|---|---|---|---|---|---|
| | **Ave** | **Std** | **Ave** | **Std** | **Ave** | **Std** | **Ave** | **Std** |
| F1 | **0** | 0 | 2.20E-18 | 7.24E-18 | 7.72E-09 | 1.37E-08 | 2.35E-19 | 1.97E-19 |
| F2 | **9.64E-261** | 0 | 1.56E-12 | 3.35E-12 | 1.14E-05 | 8.58E-06 | 5.61E-12 | 3.08E-12 |
| F3 | **0** | 0 | 8.17E-07 | 3.72E-06 | 0.000283 | 0.000619 | 3.52E+01 | 2.53E+01 |
| F4 | **5.93E-233** | 0 | 4.93E-06 | 1.11E-05 | 0.165671 | 0.080876 | 3.35E-03 | 1.30E-03 |
| F5 | **5.873546** | 0.634312 | 8.930067 | 0.171717 | 28.25196 | 62.01179 | 8.033931 | 5.981155 |
| F6 | 5.42E-06 | 2.89E-06 | 0.218349 | 0.203071 | 1.335225 | 0.404781 | **2.05E-19** | 1.88E-19 |
| F7 | **9.04E-05** | 9.71E-05 | 0.000813 | 0.000807 | 0.032696 | 0.024015 | 0.006223 | 0.001798 |
| F8 | -3.21E+03 | 341.5937 | **-2212.45** | 77.83677 | -2730.32 | 265.2633 | -4.19E+03 | 2.16E+01 |
| F9 | **0** | 0 | 3.657066 | 4.512448 | 31.09397 | 11.43579 | 2.13E-10 | 7.15E-10 |
| F10 | **8.88E-16** | 0 | 1.93E+01 | 2.64E+00 | 3.711059 | 1.902422 | 1.87E-10 | 9.32E-11 |
| F11 | **0** | 0 | 0.070081 | 0.078762 | 0.498361 | 0.236176 | 0.001331 | 0.003798 |
| F12 | 2.60E-03 | 0.006745 | 0.037791 | 0.015315 | 2.579259 | 2.335587 | **1.40E-20** | 1.51E-20 |
| F13 | 0.082788 | 0.090575 | 0.935028 | 0.093016 | 1.513306 | 0.699265 | **4.30E-20** | 6.80E-20 |
| F14 | 1.852603 | 1.902283 | 1.324236 | 1.782733 | **1.031145** | 0.181483 | 1.392995 | 1.258379 |
| F15 | **3.08E-04** | 3.68E-08 | 0.001316 | 5.46E-05 | 0.002151 | 0.005026 | 0.001538 | 0.003609 |
| F16 | **-1.03163** | 1.09E-09 | -1.03162 | 1.32E-05 | -1.0316 | 3.31E-05 | **-1.03163** | 6.78E-16 |
| F17 | 0.304251 | 8.07E-10 | 0.304253 | 2.33E-06 | 0.30435 | 0.000331 | **0.305665** | 0.007741 |
| F18 | 3.000027 | 3.30E-05 | 3.000177 | 2.11E-04 | 3.013953 | 0.01674 | **3** | 2.22E-15 |
| F19 | -3.86173 | 2.72E-03 | -3.8546 | 0.001786 | **-3.86104** | 0.001995 | -3.86278 | 2.71E-15 |
| F20 | -3.28481 | 7.03E-02 | -2.56611 | 0.568237 | -3.24127 | 0.086369 | **-3.32199** | 5.73E-06 |
| F21 | -5.0552 | 1.23E-06 | -3.47675 | 2.009677 | -7.98884 | 2.388095 | **-9.71627** | 1.61675 |
| F22 | -5.61919 | 1.621814 | -3.84991 | 2.042037 | -9.84256 | 1.324076 | **-10.3887** | 0.071691 |
| F23 | -5.489 | 1.372025 | -4.24164 | 2.034003 | -9.11802 | 2.201779 | **-10.3576** | 0.978724 |

**Table 2.** Comparison results between DCSO and the Selected Algorithms for Modern Test Functions (IEEE CEC 2019) in terms of Average and Standard Deviation

| F | DCSO | | ChOA | | CSO | | DE | |
|---|---|---|---|---|---|---|---|---|
| | **Ave** | **Std** | **Ave** | **Std** | **Ave** | **Std** | **Ave** | **Std** |
| Cec01 | **4.09E+04** | 1.83E+03 | 4.24E+09 | 9.67E+09 | 3.57E+09 | 3.6E+09 | 1.91E+10 | 9.33E+09 |
| Cec02 | 18.34314 | 1.98E-04 | 18.40831 | 0.018587 | 19.6833 | 0.607681 | **18.34286** | 6.43E-15 |

| Cec03 | **13.7024** | 3.32E-07 | 13.70242 | 7.11E-06 | 13.70249 | 0.000454 | 13.70241 | 4.55E-06 |
|---|---|---|---|---|---|---|---|---|
| Cec04 | 55.97322 | 24.34732 | 5932.62 | 2855.207 | 244.6425 | 88.65551 | **21.26087** | 3.156334 |
| Cec05 | 2.308637 | 0.13647 | 4.209471 | 0.887398 | 2.683806 | 0.146979 | **2.175169** | 0.05137 |
| Cec06 | **6.611216** | 1.22108 | 12.15444 | 0.682671 | 11.66816 | 0.437481 | 9.244935 | 0.581621 |
| Cec07 | **207.6113** | 85.5053 | 1007.134 | 179.0166 | 462.3794 | 196.4319 | 249.976 | 104.1378 |
| Cec08 | **4.591039** | 0.713487 | 6.784621 | 0.156237 | 5.967812 | 0.414544 | 5.307605 | 0.456854 |
| Cec09 | 5.089876 | 0.990137 | 449.2725 | 245.4902 | 11.81882 | 7.975786 | **3.490228** | 0.046099 |
| Cec10 | **20.48603** | 3.006093 | 21.49854 | 0.071956 | 21.43583 | 0.086047 | 21.09283 | 0.51391 |

Figure 6 presents the ranking of algorithms in the experiment, in which DCSO algorithm ranks first in the comparison (see Table 9 in the appendix for the ranking details). However, these results only show the overall performance of the proposed algorithm. Therefore, statistical tests are also required to prove the significance of the results. Thus, Wilcoxon sum rank tests were conducted to calculate P-values and the results are presented in Table 3. In all functions, except for F5, where DCSO ranks better, P-values are less than 0.05, which proves the significance of the results. Hence, the null hypothesis is rejected, as there is no difference between the means.

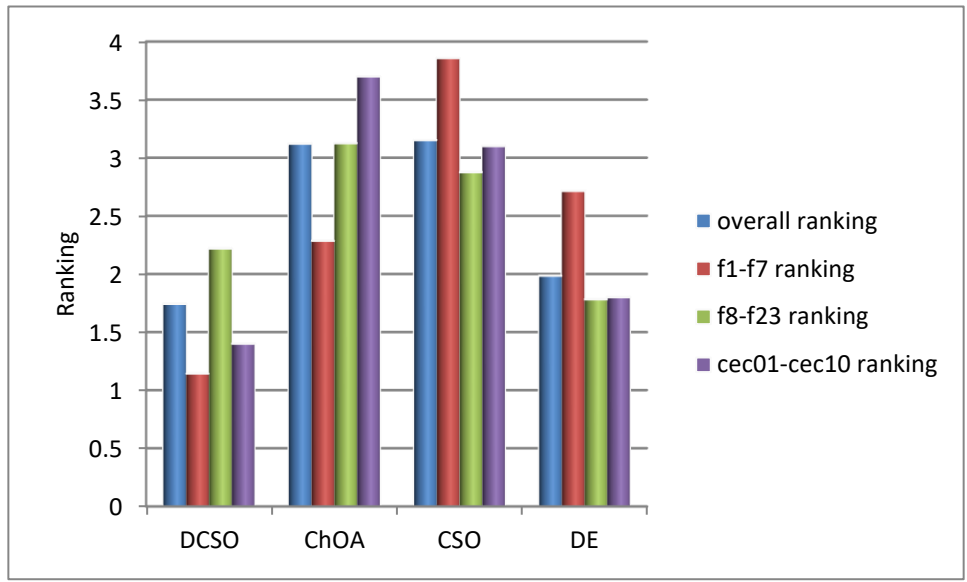

**Figure 6.** Ranking of the selected algorithms in the experiment for all the test functions using Friedman test

**Table 3.** Wilcoxon Sum Rank Tests between DCSO and the selected algorithms for all test functions

| Functions | DCSO vs. ChOA (P-values) | DCSO vs. CSO (P-values) | DCSO vs. DE (P-values) | Functions | DCSO vs. ChOA (P-values) | DCSO vs. CSO (P-values) | DCSO vs. DE (P-values) |
|---|---|---|---|---|---|---|---|
| F1 | 1.211e-12 | 1.211e-12 | 1.210e-12 | F18 | 0.000001 | 6.764e-17 | 1.211e-12 |
| F2 | 1.691e-17 | 3.017e-11 | 1.691e-17 | F19 | 9.895e-11 | 8.186e-07 | 1.207e-12 |
| F3 | 1.211e-12 | 1.211e-12 | 1.211e-12 | F20 | 3.297e-15 | 0.000004 | 3.903e-10 |
| F4 | 3.017e-11 | 1.691e-17 | 3.016e-11 | F21 | 1.691e-17 | 0.000275 | 4.329e-10 |
| F5 | 3.017e-11 | 1.691e-17 | 0.797323 | F22 | 2.493e-13 | 0.000002 | 2.212e-10 |
| F6 | 1.691e-17 | 1.691e-17 | 1.691e-17 | F23 | 3.121e-13 | 0.000143 | 2.364e-11 |
| F7 | 1.674e-12 | 1.691e-17 | 1.691e-17 | **Cec01** | 1.691e-17 | 1.691e-17 | 3.01e-11 |
| F8 | 1.691e-17 | 2.101e-07 | 1.720e-12 | **Cec02** | 1.691e-17 | 1.691e-17 | 1.144e-12 |
| F9 | 5.772e-11 | 1.211e-12 | 4.573e-12 | Cec03 | 6.474e-12 | 1.513e-11 | 9.405e-12 |
| F10 | 4.566e-13 | 1.211e-12 | 1.207e-12 | Cec04 | 1.691e-17 | 3.213e-16 | 8.487e-10 |
| F11 | 4.573e-12 | 1.211e-12 | 1.211e-12 | Cec05 | 1.691e-17 | 3.631e-12 | 0.000003 |
| F12 | 6.691e-11 | 4.501e-11 | 3.016e-11 | Cec06 | 1.691e-17 | 1.691e-17 | 3.011e-10 |
| F13 | 3.017e-11 | 3.017e-11 | 3.017e-11 | Cec07 | 1.691e-17 | 5.868e-09 | 0.062485 |
| F14 | 0.004449 | 0.015730 | 0.115737 | Cec08 | 1.691e-17 | 3.893e-13 | 0.000045 |
| F15 | 2.985e-11 | 2.985e-11 | 2.984e-11 | Cec09 | 1.691e-17 | 5.868e-09 | 2.871e-11 |
| F16 | 9.309e-12 | 9.309e-12 | 0.002773 | Cec10 | 1.691e-17 | 1.691e-17 | 4.373e-09 |
| F17 | 1.402e-11 | 1.402e-11 | 0.004449 | | | | |

## 5.3 Results and discussions for the Backboard Wiring Problem

Dataset instances for the Backboard Wiring Problem are taken from QAPLIB [15]. The dimensions of the instances are 36. There are a number of feasible choices for the distance matrix $D = (d_{jl})$; in this paper Steinberg used 1-norm, 2-norm, and squared 2-norm distances between the backboard locations which are known as ste36a, ste36b and ste36c respectively [14]. In the experiment, the DCSO algorithm is compared with Chimp Optimization Algorithm (ChOA) [11], Cat Swarm Optimization algorithm (CSO) [12] and Differential Evolution (DE) [13]. For each dataset instance, the selected algorithms are executed for 30 independent runs. For each time, the number of search agents and iterations is equal to 30 and 500 respectively. The parameter settings for the algorithms are presented in Table 8 in the appendix. The results are presented in Table 4 in terms of mean and standard deviation. Moreover, elapsed time measurement is also added to the table to achieve a fair comparison [20]. Furthermore, Figures 7, 8 and 9 shows the convergence curve of the selected algorithms for these datasets. The convergence curve for an algorithm shows the best-obtained cost values over the course of 500 iterations.

It can be noticed that DCSO algorithm has a better and smoother convergence compared to other algorithms. Furthermore, the results which are presented in Table 4 show that the DCSO algorithm returns the least cost values for all the datasets and outperforms the other competitive algorithms. Regarding the elapsed time measurement, the DE algorithm takes the least amount of time compared to the rest. DCSO takes double amount of time in comparison with the DE algorithm. However, figures 7, 8, and 9 show that the DCSO achieve better results in almost one third of the iterations. As a result, we can conclude that even though the DCSO requires more computation time per an iteration, it is more efficient than the DE algorithm on the whole. Finally, it is worth mentioning that the simulation was conducted using MATLAB on an Intel(R) Core(TM) i7-2670 with 6GB RAM. Therefore, different hardware specifications would surely yield unlike elapsed time results.

**Table 4.** Comparison results of the selected algorithms for backboard wiring problem (QAPLIB) in terms of average, standard deviation and elapsed time

| Algorithms | | Ste36a | Ste36b | Ste36c |
|---|---|---|---|---|
| **DCSO** | Ave | **13431.73** | **29300.27** | **10700541** |
| | Std | 916.932 | 2832.584 | 586778.3 |
| | Elapsed time (s) | 3.963525 | 3.926989 | 3.961443 |
| **ChOA** | Ave | 15073.93 | 42990.93 | 12707350 |
| | Std | 454.0811 | 4174.001 | 362543.5 |
| | Elapsed time (s) | 11.624100 | 11.493323 | 11.484724 |
| **CSO** | Ave | 15230.8 | 41188.13 | 12750790 |
| | Std | 564.1728 | 2511.199 | 457585.4 |
| | Elapsed time (s) | 7.682724 | 7.657057 | 7.634772 |
| **DE** | Ave | 14100.53 | 36701.2 | 11915816 |
| | Std | 397.7436 | 1790.165 | 313946.6 |
| | Elapsed time (s) | **2.099774** | **2.063565** | **2.065695** |

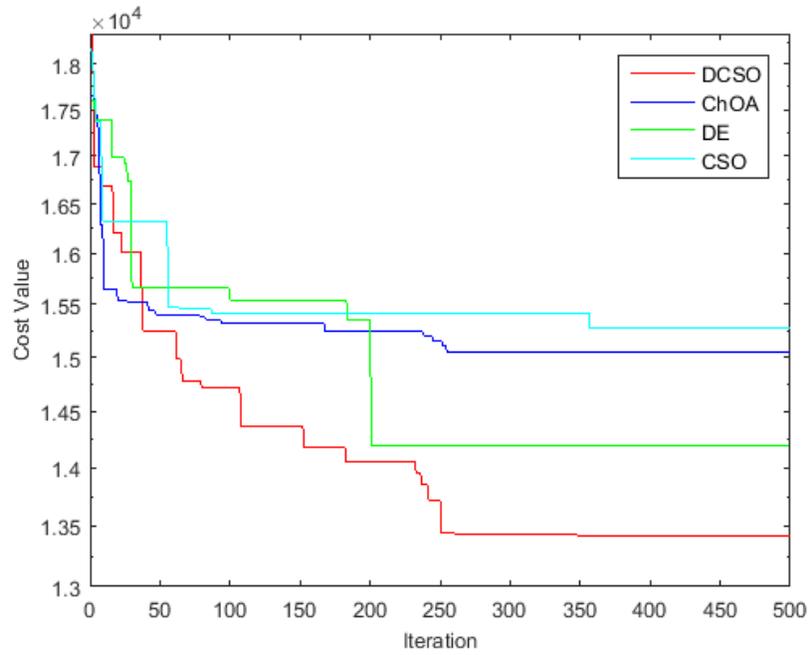

**Figure 7.** Convergence curve of DCSO and its competitive algorithms for Ste36a dataset

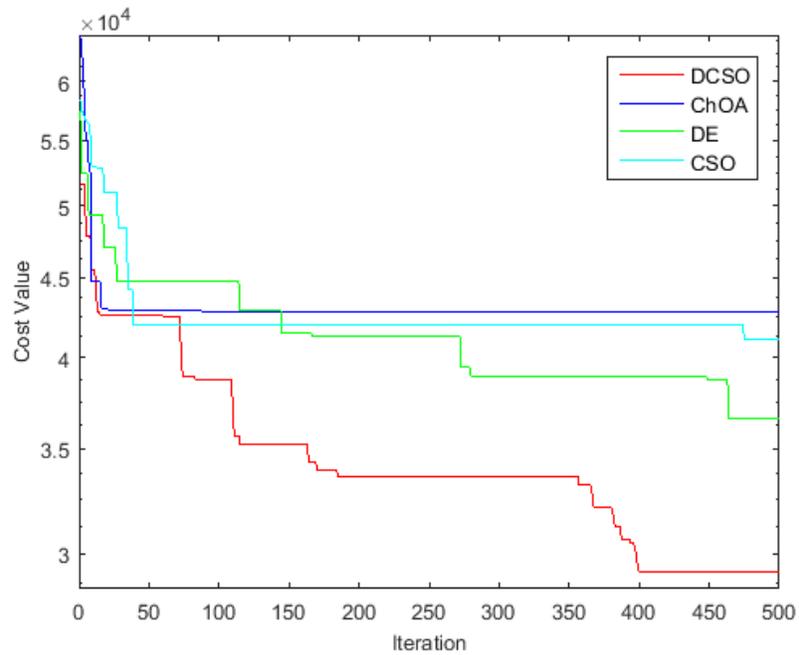

**Figure 8.** Convergence curve of DCSO and its competitive algorithms for Ste36b dataset

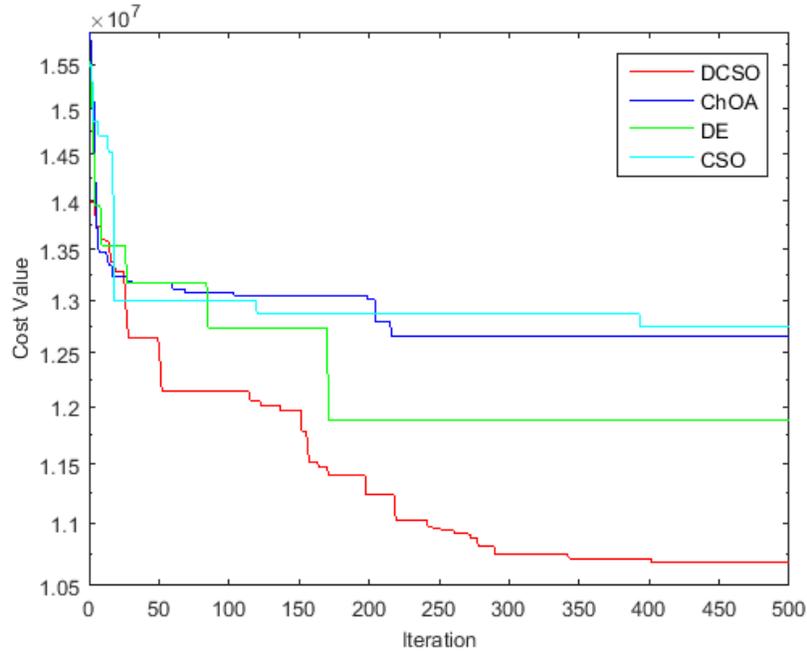

**Figure 9.** Convergence curve of DCSO and its competitive algorithms for Ste36c dataset

5.4 Exploration and Exploitation measurements

Swarm individuals generally have two basic search behaviors, which are known as exploration and exploitation. In the first one, the individuals are diverging and the distances between their dimensions are increasing. This phase is used to discover new areas and escape from possible local optima trap. On the other hand, in the exploitation phase, the individuals are intensifying and the distance between their dimensions are decreasing. In this phase, the individuals tend to locally search in the neighborhood and converge towards the global optimum. Finding a proper balance between the exploration and exploitation phases is a key factor to prevent premature convergence in metaheuristic algorithms [21].

In this study, the dimension-wise diversity measurement [21] is adopted to quantitatively measure the exploration and exploitation degree of the selected algorithms. Also, this study substituted mean by median in equation (12) because it represents the center of the population more accurately [22], [23].

$$Div_j = \frac{1}{n}\sum_{i=1}^{n} meadian\,(x^j) - x_i^j; \qquad (12)$$

$$Div = \frac{1}{D}\sum_{j=1}^{D} Div_j \qquad (13)$$

Where median ($x^j$) depicts the median of dimension j in the entire population. $x_i^j$ is the dimension j of search agent i. n is the population size.

The diversity in each dimension $Div_j$ is formulated as the average distance between the dimension j of every search agent and the median of that dimension. Then, the average of diversity for the whole dimensions is computed in Div. The percentage of exploration and exploitation in an algorithm can be calculated by taking the average of the following:

$$XPL\% = \left(\frac{Div}{Div_{max}}\right) * 100 \tag{14}$$

$$XPT\% = \left(\frac{|Div - Div_{max}|}{Div_{max}}\right) * 100 \tag{15}$$

Where $Div_{max}$ is the maximum diversity value found in the whole optimization process. XPL% and XPT% are the degree of exploration and exploitation respectively and they complement each other. Table 5 presents the explorative and exploitative capabilities of DCSO and the selected algorithms for all test functions and datasets used in the experiment. For each test function or dataset, each algorithm is executed for 30 independent runs and the average values of these 30 runs are presented in Table 5. In each run of the experiment, the number of search agents and iterations are equal to 30 and 500 respectively.

Furthermore, by looking at Table 5, it can be noticed that the algorithms do not have the same percentage of exploration and exploitation for the different test functions or datasets. This means that this measurement, to some degree, is problem-dependent. However, on average, the ratio between these two phases in DCSO is close to 50%:50%, while this ratio for CSO is approximately 75%:25%. Therefore, it can be concluded that DCSO algorithm is considerably more balanced as compared to the original CSO algorithm. Regarding ChOA and DE algorithms, they have an explorative nature except for ChOA algorithm in which it becomes exploitative for the QAP datasets.

In the case of DCSO and CSO algorithms approaching the balance ratio of 50%:50% provided better results. However, this might not be true for every algorithm. The reason is that there are other factors that may play crucial roles in metaheuristic algorithms such as the structure of the algorithm, the relationship and coherence between the search agents, parameter setting, etc.

**Table 5.** Exploration and exploitation percentages of the selected algorithms for all the benchmark functions and the QAP datasets

| Functions/ datasets | DCSO | | ChOA | | CSO | | DE | |
|---|---|---|---|---|---|---|---|---|
| | Average XPL% | Average XPT% | Average XPL% | Average XPT% | Average XPL% | Average XPT% | Average XPL% | Average XPT% |
| F1 | 41.07976 | 58.92024 | 83.95463 | 16.04537 | 79.38937 | 20.61063 | 74.56291 | 25.43709 |
| F2 | 42.40773 | 57.59226 | 82.68069 | 17.31931 | 78.61779 | 21.38221 | 69.51993 | 30.48007 |
| F3 | 39.15998 | 60.84001 | 80.15966 | 19.84034 | 76.8929 | 23.1071 | 72.88248 | 27.11752 |
| F4 | 36.50502 | 63.49497 | 83.71753 | 16.28247 | 78.83882 | 21.16118 | 84.60692 | 15.39308 |
| F5 | 45.98909 | 54.01090 | 81.3323 | 18.6677 | 79.39371 | 20.60629 | 77.03315 | 22.96685 |
| F6 | 54.76994 | 45.23005 | 84.66168 | 15.33832 | 78.95073 | 21.04927 | 74.60318 | 25.39682 |
| F7 | 51.08535 | 48.91464 | 84.11816 | 15.88184 | 75.30637 | 24.69363 | 86.6364 | 13.3636 |
| F8 | 58.26718 | 41.73281 | 85.39696 | 14.60304 | 87.07155 | 12.92845 | 24.41982 | 75.58018 |
| F9 | 30.89491 | 69.10508 | 81.64988 | 18.35012 | 74.91016 | 25.08984 | 76.6011 | 23.3989 |
| F10 | 38.06791 | 61.93208 | 84.80917 | 15.19083 | 77.60153 | 22.39847 | 73.41518 | 26.58482 |
| F11 | 31.68670 | 68.31329 | 82.24892 | 17.75108 | 79.4591 | 20.5409 | 74.84748 | 25.15252 |
| F12 | 57.84533 | 42.15466 | 85.58024 | 14.41976 | 78.42027 | 21.57973 | 73.56594 | 26.43406 |
| F13 | 47.12821 | 52.87178 | 80.1395 | 19.8605 | 80.99657 | 19.00343 | 17.39646 | 82.60354 |
| F14 | 50.79708 | 49.20291 | 75.44297 | 24.55703 | 88.15168 | 11.84832 | 66.04839 | 33.95161 |
| F15 | 48.36300 | 51.63699 | 77.09498 | 22.90502 | 75.17235 | 24.82765 | 47.58012 | 52.41988 |
| F16 | 52.60580 | 47.39419 | 65.53101 | 34.46899 | 77.54581 | 22.45419 | 38.5467 | 61.4533 |
| F17 | 55.48436 | 44.51563 | 75.64292 | 24.35708 | 93.46458 | 6.535416 | 34.87629 | 65.12371 |
| F18 | 54.2772 | 45.72272 | 72.70525 | 27.29475 | 75.62741 | 24.37259 | 66.12894 | 33.87106 |
| F19 | 42.2657 | 57.73428 | 54.61856 | 45.38144 | 58.72203 | 41.27797 | 3.177761 | 96.82224 |
| F20 | 47.92797 | 52.07202 | 57.55077 | 42.44923 | 51.75539 | 48.24461 | 60.75544 | 39.24456 |
| F21 | 40.46087 | 59.53912 | 61.73172 | 38.26828 | 54.27163 | 45.72837 | 50.16661 | 49.83339 |
| F22 | 41.34661 | 58.65338 | 61.75452 | 38.24548 | 54.639 | 45.361 | 39.93836 | 60.06164 |
| F23 | 39.8822 | 60.1177 | 64.08032 | 35.91968 | 53.77755 | 46.22245 | 42.31111 | 57.68889 |
| CEC01 | 51.34380 | 48.65619 | 80.62148 | 19.37852 | 75.14055 | 24.85945 | 91.79923 | 8.200775 |
| CEC02 | 55.15344 | 44.84655 | 86.89284 | 13.10716 | 76.92232 | 23.07768 | 89.61158 | 10.38842 |
| CEC03 | 55.51488 | 44.48511 | 86.99009 | 13.00991 | 74.3631 | 25.6369 | 94.47115 | 5.528848 |
| CEC04 | 62.21963 | 37.78036 | 80.28436 | 19.71564 | 69.88884 | 30.11116 | 42.91872 | 57.08128 |
| CEC05 | 60.25675 | 39.74324 | 79.77446 | 20.22554 | 73.68166 | 26.31834 | 45.30146 | 54.69854 |
| CEC06 | 43.74468 | 56.25531 | 85.1097 | 14.8903 | 88.64608 | 11.35392 | 95.47904 | 4.52096 |
| CEC07 | 51.50923 | 48.49076 | 84.87551 | 15.12449 | 86.39797 | 13.60203 | 94.38941 | 5.610588 |
| CEC08 | 51.71601 | 48.28398 | 85.22845 | 14.77155 | 87.86933 | 12.13067 | 93.93573 | 6.064269 |
| CEC09 | 67.90956 | 32.09043 | 83.94941 | 16.05059 | 75.34462 | 24.65538 | 62.4828 | 37.5172 |
| CEC10 | 42.23906 | 57.76093 | 84.91573 | 15.08427 | 89.51355 | 10.48645 | 96.45494 | 3.545056 |
| Ste36a | 32.02769 | 67.97230 | 14.78834 | 85.21166 | 80.59219 | 19.40781 | 97.61903 | 2.380971 |
| Ste36b | 33.7079 | 66.2920 | 21.59497 | 78.40503 | 81.53616 | 18.46384 | 97.81146 | 2.188536 |
| Ste36c | 35.11633 | 64.88367 | 17.38805 | 82.61195 | 80.05343 | 19.94657 | 97.63377 | 2.366231 |

## 6. Conclusion and Future Directions

CSO algorithm is hugely limited by the shortcoming of premature convergence, which is the likelihood of falling into local optima. Balancing the exploration and exploitation phases can be used to address this issue. Therefore, in this paper, DCSO algorithm was proposed that modifies the selection scheme and the seeking mode of the existing CSO algorithm. These modifications gave a dynamic nature to the algorithm and more importantly provided a proper balance between the exploration and exploitation phases of the algorithm. Moreover, to confirm this balance and expose the percentage of these phases the Dimension-wise diversity measurement was also used. The results show on average, the proportion between these two phases in DCSO is close to 50%:50%, while this ratio for CSO is roughly 75%:25%. In the experiment, the robustness of the proposed algorithm was validated by testing it on a total of 33 benchmark functions and a real world scenario called backboard wiring problem. The results are then compared with three other well-known algorithms. We can conclude that the proposed method addresses the premature convergence issue and yields very competitive results. Having said that, there is still room for further enhancements in the algorithm; For example, combining the seeking mode of the algorithm with a suitable local search technique such as Golden Section search might considerably enhance the efficiency of the algorithm.

### Data availability
The MATLAB code for DCSO algorithm can be found in the below repository link: https://github.com/aramahmed/DCSO-Algorithm/

### Conflict of Interest
The authors declare that they have no conflict of interest.

## References

[1] Baskan, O. (Ed.). (2016). Optimization Algorithms: Methods and Applications. BoD–Books on Demand.
[2] Xin-She, Y. (2010). An introduction with metaheuristic applications. Engineering Optimization. John Wiley, New York, NY, USA
[3] Voß, S. (2000, August). Meta-heuristics: The state of the art. In Workshop on Local Search for Planning and Scheduling (pp. 1-23). Springer, Berlin, Heidelberg.
[4] Yang, X. S. (2020). Nature-inspired optimization algorithms: challenges and open problems. Journal of Computational Science, 101104.

[5] Yang, X. S. (2010). *Nature-inspired metaheuristic algorithms*. Luniver press.
[6] Pappula, L., & Ghosh, D. (2018). Cat swarm optimization with normal mutation for fast convergence of multimodal functions. *Applied soft computing*, *66*, 473-491.
[7] Nie, X., Wang, W., & Nie, H. (2017). Chaos quantum-behaved cat swarm optimization algorithm and its application in the PV MPPT. *Computational intelligence and neuroscience*.
[8] Orouskhani, M., Mansouri, M., & Teshnehlab, M. (2011, June). Average-inertia weighted cat swarm optimization. In *International conference in swarm intelligence* (pp. 321-328). Springer, Berlin, Heidelberg.
[9] Ahmed, A. M., Rashid, T. A., & Saeed, S. A. M. (2020). Cat Swarm Optimization Algorithm: A Survey and Performance Evaluation. *Computational Intelligence and Neuroscience*, *2020*.
[10] Abdel-Basset, M., Abdel-Fatah, L., & Sangaiah, A. K. (2018). Metaheuristic algorithms: A comprehensive review. In *Computational intelligence for multimedia big data on the cloud with engineering applications* (pp. 185-231). Academic Press.
[11] Khishe, M., & Mosavi, M. R. (2020). Chimp optimization algorithm. *Expert Systems with Applications*, 113338.
[12] Chu, S. C., & Tsai, P. W. (2007). Computational intelligence based on the behavior of cats. *International Journal of Innovative Computing, Information and Control*, *3*(1), 163-173.
[13] Storn, R., & Price, K. (1997). Differential evolution–a simple and efficient heuristic for global optimization over continuous spaces. *Journal of global optimization*, *11*(4), 341-359.
[14] Steinberg, L. (1961). The backboard wiring problem: A placement algorithm. Siam Review, 3(1), 37-50.
[15] Burkard, R. E., Karisch, S. E., & Rendl, F. (1997). QAPLIB–a quadratic assignment problem library. *Journal of Global optimization*, *10*(4), 391-403. Accessed: 27 June 2020
[16] Tosun, U. (2015). On the performance of parallel hybrid algorithms for the solution of the quadratic assignment problem. *Engineering applications of artificial intelligence*, *39*, 267-278.
[17] Fard, A. M. F., & Hajiaghaei-Keshteli, M. (2018). A bi-objective partial interdiction problem considering different defensive systems with capacity expansion of facilities under imminent attacks. *Applied Soft Computing*, *68*, 343-359.
[18] Chu, S. C., Tsai, P. W., & Pan, J. S. (2006, August). Cat swarm optimization. In *Pacific Rim international conference on artificial intelligence* (pp. 854-858). Springer, Berlin, Heidelberg.
[19] Price, K. V., Awad, N. H., Ali, M. Z., & Suganthan, P. N. (2018). The 100-digit challenge: Problem definitions and evaluation criteria for the 100-digit challenge special session and competition on single objective numerical optimization. *Nanyang Technological University,, Singapore*.
[20] Nguyen, T. T., Vo, D. N., & Dinh, B. H. (2018). An effectively adaptive selective cuckoo search algorithm for solving three complicated short-term hydrothermal scheduling problems. *Energy*, *155*, 930-956.
[21] Cheng, S., Shi, Y., Qin, Q., Zhang, Q., & Bai, R. (2014). Population diversity maintenance in brain storm optimization algorithm. *Journal of Artificial Intelligence and Soft Computing Research*, *4*(2), 83-97.
[22] Hussain, K., Salleh, M. N. M., Cheng, S., & Shi, Y. (2019). On the exploration and exploitation in popular swarm-based metaheuristic algorithms. *Neural Computing and Applications*, *31*(11), 7665-7683.
[23] Morales-Castañeda, B., Zaldivar, D., Cuevas, E., Fausto, F., & Rodríguez, A. (2020). A better balance in metaheuristic algorithms: Does it exist?. *Swarm and Evolutionary Computation*, 100671.
[24] Mirjalili, S., & Lewis, A. (2016). The whale optimization algorithm. *Advances in engineering software*, *95*, 51-67.

# Appendix

**Table 6.** Details of the Test Functions that are used in the experiments [24]

| FUNCTIONS | DIMENSION | RANGE | $f_{min}$ |
|---|---|---|---|
| $F_1(x) = \sum_{i=1}^{N} x_i^2$ | 30 | [-100, 100] | 0 |
| $F_2(x) = \sum_{i=1}^{N} |x_i| + \prod_{i=1}^{N} |x_i|$ | 30 | [-10, 10] | 0 |
| $F_3(x) = \sum_{i=1}^{N} \left(\sum_{j-1}^{i} x_j\right)^2$ | 30 | [-100, 100] | 0 |
| $F_4(x) = \max_i\{|x_i|, \quad 1 \le i \le N\}$ | 30 | [-100, 100] | 0 |
| $F_5(x) = \sum_{i-1}^{N-1} [100(x_{i+1} - x_i^2)^2 + (x_i - 1)^2]$ | 30 | [-30, 30] | 0 |
| $F_6(x) = \sum_{i=1}^{N} ([x_i + 0.5])^2$ | 30 | [-100, 100] | 0 |
| $F_7(x) = \sum_{i=1}^{N} ix_i^4 + \text{random}[0,1]$ | 30 | [-1.28, 1.28] | 0 |
| $F_8(x) = \sum_{i=1}^{N} -x_i \sin\left(\sqrt{|x_i|}\right)$ | 30 | [-500, 500] | -418.9829×5 |
| $F_9(x) = \sum_{i=1}^{N} [x_i^2 - 10\cos(2\pi x_i) + 10]$ | 30 | [-5.12, 5.12] | 0 |
| $F_{10}(x) = -20\exp\left(-0.2\sqrt{\frac{1}{N}\sum_{i=1}^{N} x_i^2}\right) - \exp\left(\frac{1}{N}\sum_{i=1}^{N} \cos(2\pi x_i)\right) + 20 + e$ | 30 | [-32, 32] | 0 |
| $F_{11}(x) = \frac{1}{4000}\sum_{i=1}^{N} x_i^2 - \prod_{i=1}^{N} \cos\left(\frac{x_i}{\sqrt{i}}\right) + 1$ | 30 | [-600, 600] | 0 |
| $F_{12}(x) = \frac{\pi}{N}\{10\sin(\pi y_1) + \sum_{i=1}^{N-1}(y_i - 1)^2 [1 + 10\sin^2(\pi y_{i+1})] + (y_N - 1)^2\} + \sum_{i=1}^{N} u(x_i, 10, 100, 4)$ $y_i = 1 + \frac{x_i + 1}{4} \quad u(x_i, a, k, m) = \begin{cases} k(x_i - a)^m & x_i > a \\ 0 & -a < x_i < a \\ k(-x_i - a)^m & x_i < -a \end{cases}$ | 30 | [-50, 50] | 0 |
| $F_{13}(x) = 0.1\{\sin^2(3\pi x_1) + \sum_{i=1}^{N}(x_i - 1)^2 [1 + \sin^2(3\pi x_i + 1)] + (x_N - 1)^2 [1 + \sin^2(2\pi x_N)]\} + \sum_{i=1}^{N} u(x_i, 10, 100, 4)$ | 30 | [-50, 50] | 0 |
| $F_{14}(x) = \left(\frac{1}{500} + \sum_{j=1}^{25} \frac{1}{j + \sum_{i=1}^{2}(x_i - a_{ij})^6}\right)^{-1}$ | 2 | [-65, 65] | 1 |
| $F_{15}(x) = \sum_{i=1}^{11} [a_i - \frac{x_1(b_i^2 + b_i x_2)}{b_i^2 + b_i x_3 + x_4}]^2$ | 4 | [-5, 5] | 0.00030 |
| $F_{16}(x) = 4x_1^2 - 2.1x_1^4 + \frac{1}{3}x_1^6 + x_1 x_2 - 4x_2^2 + 4x_2^4$ | 2 | [-5, 5] | -1.398 |
| $F_{17}(x) = \left(x_2 - \frac{5.1}{4\pi^2}x_1^2 + \frac{5}{\pi}x_1 - 6\right)^2 + 10\left(1 - \frac{1}{8\pi}\right)\cos x_1 + 10$ | 2 | [-5, 5] | 0.398 |
| $F_{18}(x) = [1 + (x_1 + x_2 + 1)^2(19 - 14x_1 + 3x_1^2 - 14x_2 + 6x_1 x_2 + 3x_2^2)]$ $\times [30 + (2x_1 - 3x_2)^2 \times (18 - 32x_1 + 12x_1^2 + 48x_2 - 36x_1 x_2 + 27x_2^2)]$ | 2 | [-2, 2] | 3 |
| $F_{19}(X) = -\sum_{I=1}^{4} C_I \exp\left(-\sum_{J=1}^{3} A_{IJ}(X_J - P_{IJ})^2\right)$ | 3 | [1, 3] | -3.86 |
| $F_{20}(X) = -\sum_{I=1}^{4} C_I \exp\left(-\sum_{J=1}^{6} A_{IJ}(X_J - P_{IJ})^2\right)$ | 6 | [0, 1] | -3.32 |
| $F_{21}(X) = -\sum_{I=1}^{5} [(X - A_I)(X - A_I)^T + C_I]^{-1}$ | 4 | [0, 10] | -10.1532 |
| $F_{22}(X) = -\sum_{I=1}^{7} [(X - A_I)(X - A_I)^T + C_I]^{-1}$ | 4 | [0, 10] | -10.4028 |
| $F_{23}(X) = -\sum_{I=1}^{10} [(X - A_I)(X - A_I)^T + C_I]^{-1}$ | 4 | [0, 10] | -10.5363 |

**Table 7.** CEC 2019 Benchmarks "the 100-digit challenge". Refer to [19] for more details

| No. | Functions | Dimension | Range | $f_{min}$ |
|---|---|---|---|---|
| CEC01 | Storn's Chebyshev Polynomial Fitting Problem | 9 | [-8192, 8192] | 1 |
| CEC02 | Inverse Hilbert Matrix Problem | 16 | [-16384, 16384] | 1 |
| CEC03 | Lennard-Jones Minimum Energy Cluster | 18 | [-4, 4] | 1 |
| CEC04 | Rastrigin's Function | 10 | [-100, 100] | 1 |
| CEC05 | Griewangk's Function | 10 | [-100, 100] | 1 |
| CEC06 | Weierstrass Function | 10 | [-100, 100] | 1 |
| CEC07 | Modified Schwefel's Function | 10 | [-100, 100] | 1 |
| CEC08 | Expanded Schaffer's F6 Function | 10 | [-100, 100] | 1 |
| CEC09 | Happy Cat Function | 10 | [-100, 100] | 1 |
| CEC10 | Ackley Function | 10 | [-100, 100] | 1 |

**Table 8.** Control parameter values for the selected algorithms

| Algorithms | Parameters | Values |
|---|---|---|
| DCSO | Population Size | 30 |
| | Max Iteration | 500 |
| | Inertia Weight W(I) | [0.4, 0.9] |
| | SMP | 5 |
| | CDC | 0.8% |
| | Constant (C) | 2.05 |
| | Random Values | [0, 1] |
| ChOA | Population Size | 30 |
| | Max Iteration | 500 |
| | Random Values: r1, r2 | [0, 1] |
| | M | Chaotic |
| CSO | Population Size | 30 |
| | Max Iteration | 500 |
| | Constant (C) | 2.05 |
| | Random Values | [0, 1] |
| | SMP | 5 |
| | CDC | 0.8 |
| | MR | 0.2 |
| | SRD | 0.2 |
| | CONSTANT (C) | 2 |
| | POPULATION SIZE | 30 |
| DE | MAX ITERATION | 500 |
| | BETA_MIN | 0.2 |
| | BETA_MAX | 0.8 |
| | CROSSOVER RATE | 0.2 |

**Table 9.** Ranking of the selected algorithms for all test functions (Friedman test)

| TF | Ranking DCSO | Ranking ChOA | Ranking CSO | Ranking DE |
|---|---|---|---|---|
| F1 | 1 | 2 | 4 | 3 |
| F2 | 1 | 2 | 4 | 3 |
| F3 | 1 | 2 | 3 | 4 |
| F4 | 1 | 2 | 4 | 3 |
| F5 | 1 | 3 | 4 | 2 |
| F6 | 2 | 3 | 4 | 1 |
| F7 | 1 | 2 | 4 | 3 |
| F8 | 3 | 1 | 2 | 4 |
| F9 | 1 | 3 | 4 | 2 |
| F10 | 1 | 4 | 3 | 2 |
| F11 | 1 | 3 | 4 | 2 |
| F12 | 2 | 3 | 4 | 1 |
| F13 | 2 | 3 | 4 | 1 |
| F14 | 4 | 2 | 1 | 3 |
| F15 | 1 | 2 | 4 | 3 |
| F16 | 1.5 | 3 | 4 | 1.5 |
| F17 | 4 | 3 | 2 | 1 |
| F18 | 2 | 3 | 4 | 1 |
| F19 | 2 | 4 | 1 | 3 |
| F20 | 2 | 4 | 3 | 1 |
| F21 | 3 | 4 | 2 | 1 |
| F22 | 3 | 4 | 2 | 1 |
| F23 | 3 | 4 | 2 | 1 |
| Cec01 | 1 | 3 | 2 | 4 |
| Cec02 | 2 | 3 | 4 | 1 |
| Cec03 | 1 | 3 | 4 | 2 |
| Cec04 | 2 | 4 | 3 | 1 |
| Cec05 | 2 | 4 | 3 | 1 |
| Cec06 | 1 | 4 | 3 | 2 |
| Cec07 | 1 | 4 | 3 | 2 |
| Cec08 | 1 | 4 | 3 | 2 |
| Cec09 | 2 | 4 | 3 | 1 |
| Cec10 | 1 | 4 | 3 | 2 |
| F1-F7 Subtotal | 8 | 16 | 27 | 19 |
| F1-F7 Ranking | 1.142857 | 2.285714 | 3.857143 | 2.714286 |
| F8-F23 Subtotal | 35.5 | 50 | 46 | 28.5 |
| F8-F23 Ranking | 2.21875 | 3.125 | 2.875 | 1.78125 |
| CEC01-CEC10 Subtotal | 14 | 37 | 31 | 18 |
| CEC01-CEC10 Ranking | 1.4 | 3.7 | 3.1 | 1.8 |
| Total | 57.5 | 103 | 104 | 65.5 |
| Overall ranking | 1.742424 | 3.121212 | 3.151515 | 1.984848 |